# TECHNICAL PROGRESS ANALYSIS USING A DYNAMIC TOPIC MODEL FOR TECHNICAL TERMS TO REVISE PATENT CLASSIFICATION CODES

## M.IWATA[1] , Y.MATSUDA[2] , Y.UTSUMI[3] , Y.TANAKA[4] , K.NAKATA[5]

[1]Department of Industrial Engineering and Economics
Tokyo Institute of Technology, Tokyo, Japan
iwata.m.ac@m.titech.ac.jp

[2]Intellectual Property Department
Rakuten,Inc., Tokyo, Japan
yoshiro.matsuda@rakuten.com

[3]Intellectual Property Department
Rakuten,Inc., Tokyo, Japan
yoshimasa.utsumi@rakuten.com

[4]Department of Industrial Engineering and Economics
Tokyo Institute of Technology, Tokyo, Japan
tanaka.y.al@m.titech.ac.jp

[5]Department of Industrial Engineering and Economics
Tokyo Institute of Technology, Tokyo, Japan
nakata.k.ac@m.titech.ac.jp

## ABSTRACT

Japanese patents are assigned a patent classification code, FI (File Index), that is unique to Japan. FI is a subdivision of the IPC, an international patent classification code, that is related to Japanese technology. FIs are revised to keep up with technological developments. These revisions have already established more than 30,000 new FIs since 2006. However, these revisions require a lot of time and workload. Moreover, these revisions are not automated and are thus inefficient. Therefore, using machine learning to assist in the revision of patent classification codes (FI) will lead to improved accuracy and efficiency. This study analyzes patent documents from this new perspective of assisting in the revision of patent classification codes with machine learning.

To analyze time-series changes in patents, we used the dynamic topic model (DTM), which is an extension of the latent Dirichlet allocation (LDA). Also, unlike English, the Japanese language requires morphological analysis. Patents contain many technical words that are not used in everyday life, so morphological analysis using a common dictionary is not sufficient. Therefore, we used a technique for extracting technical terms from text. After extracting technical terms, we applied them to DTM. In this study, we determined the technological progress of the lighting class F21 for 14 years and compared it with the actual revision of patent classification codes.

In other words, we extracted technical terms from Japanese patents and applied DTM to determine the progress of Japanese technology. Then, we analyzed the results from the new perspective of revising patent classification codes with machine learning. As a result, it was found that those whose topics were on the rise were judged to be new technologies.

## 1 INTRODUCTION

In Japan, more than 300,000 patents are applied for each year. After a patent is applied for, a patent examiner examines the patent through a search of prior literature etc. In the search, patent classification codes are used for narrowing down patents. The International Patent Classification (IPC) is an international patent classification code. However, it is unable to sufficiently narrow down the fields in which many patent applications are applied for in Japan. Therefore, Japan has introduced its own patent classification code, FI (File Index). FI is a subdivision of IPC that is related to Japanese technology. This classification code is used as a search key for domestic patent documents. Like IPC, FI has a hierarchical structure with sections, classes, subclasses, main groups, subgroups, expansion symbols, and book identification symbols. The FI consists of approximately 190,000 items. (The IPC consists of about 70,000 items.)

An example[*] of FI is below.

**F21S8/02,300・・・especially for the purpose of illuminating the feet, e.g., footlight**
F: section, F21: class, F21S: subclass, F21S8/00: main group, F21S8/02: subgroup, 300: expansion symbol

The identification symbol appears like @A and is used to divide subordinate items or development symbols of the International Patent Classification into smaller groups. In Japan, the FI is revised once or twice a year to keep up with technological developments. Under these revisions, more than 30,000 new FIs have been established since 2006, which is a huge volume of work. However, these revisions are being made by people, which is not efficient. Therefore, using machine learning to assist in the revision of patent classification codes (FI) will lead to more correct patent classification code creation and efficiency. This study identifies the progress of Japanese technology from the new perspective of assisting in the revision of patent classification codes with machine learning.

The dynamic topic model (DTM) is mainly used to analyze the time-series changes of topics in a large document. DTM is an extended model of latent Dirichlet allocation (LDA). By applying DTM to a patent document, it is possible to estimate time-series changes in technology. In addition, an analysis with DTM is conducted on the basis of the number of occurrences of words in a document. However, unlike English, Japanese does not use spaces to separate words from sentences. Therefore, for Japanese patents, it is necessary to perform the process of morphological analysis (dividing sentences into parts of speech), which is not required for English. For the Japanese language, morphological analysis using dynamic programming has been proposed. However, patents contain many technical terms (special words that are not used in everyday life), so morphological analysis using ordinary dictionaries is not sufficient. Therefore, it is necessary to have techniques for extracting technical terms from documents. In this study, we performed a morphological analysis, extracted technical terms, and applied them to DTM. We determined the technological progress for 14 years of F21, the class on lighting. We show the potential of machine learning to aid in the revision of patent classification codes.

Considering the above, the contributions in this study are:

- Proposing a new perspective of analysis in the form of revisions of patent classification codes (FI) based on machine learning.
- Extracting technical terms from Japanese patents and identifying the progress of Japanese technology by using DTM.

---

[*] https://www.j-platpat.inpit.go.jp/p1101

## 2 PREVIOUS STUDY

### 2.1 Dynamic Topic Model (DTM)

When considering the advancement of technology, the concept of "topic" is important. This is because each patent is considered to be composed of various technologies (= topics).

For the problem of estimating topics, latent Dirichlet allocation (LDA) [2] is widely used. LDA is a generative probabilistic model that extracts potential topics from discrete data, such as text data. In LDA, it is assumed that every document is composed of several topics, and each topic is based on a distribution of words. The relationships between documents and topics and between topics and words are represented as one-to-many, so the model is more realistic for text data. However, LDA does not take time series into account. Therefore, the model does not allow us to check the transitions of topics.

To address this problem, the dynamic topic model (DTM) [3] was proposed. DTM is an extension of the LDA that does take time series into account. In DTM, a dataset is sliced at specified times, and there are document topics for each time slice. Each topic in DTM is represented as a distribution of words, and the distribution of words itself changes in parallel with the change in time slice. In the paper on DTM, the model was demonstrated with the OCR'ed archives of the journal Science from 1880 to 2000. In addition, Ranaei et al. applied a dynamic topic model to patent documents to determine the emergence of the vehicle technology [4]. Thus, although not in terms of the revision of patent classification codes, a patent analysis using the topic model has been conducted.

### 2.2 Morphological Analysis

Morphological analysis is one of the fundamental techniques of natural language processing. The input text can be transformed into a set of morphemes (words) with frequencies (bag-of-words, BOW). Morphological analysis is necessary for languages where there are no spaces in sentences, such as Japanese and Chinese.

An example is given below.

**「私は、今日目黒区役所に行って書類提出を行った。」**
**⇒「私/は/、/今日/目黒/区役所/に/行っ/て/書類/提出/を/行っ/た/。」**

(Terms judged to be nouns are shown in red.)

As a result, the nouns **「私、今日、目黒、区役所、書類、提出」** were extracted.

Originally, in Japan, most of the conventional morphological analysis studies were given, a priori, only word division definitions. However, the definitions are not always optimal for individual applications, nor are they flexible enough to deal with the diversity of real-world problems. Therefore, Kudo et al. [5] proposed a method for calculating the expected value of multiple splits and for taking the approach that allows for a diversity of splitting methods. With the method, using a graph structure called a "morphological lattice," the least-cost problem is solved by dynamic programming. The algorithm is freely available in a library called MeCab[†] and is used by many people.

---
[†] https://taku910.github.io/mecab/

## 2.3 Terminology Extraction

In Japan, most of the studies on terminology extraction are based on Nakagawa et al.'s method of terminology extraction based on the frequency of occurrence and collinearity of words [6]. There is also a Python library[‡] based on this method. In [6], an approach that pays attention to sequential nouns was adopted. Papers on the extraction of patent terminology also basically adopt this idea. Some sites[§] actually adopted this paper and applied it to patent data. However, the work of Nakagawa et al. [6] is not limited to the field of patents. Yuzukiyama et al. [7] adopted the idea of [6] in the extraction of technical terms related to the patent field and also showed that the idea of [6] is effective in the patent field. In this study, no consideration was given to patent-specific expressions. In the discussion of this study, the author suggests that patent-specific terms should be excluded. Specifically, there were the following two points.

- Patent-specific terms such as "**該当,次,前記**" should be excluded.
- Symbols such as ABC, colors, and numbers may be noise in documents.

Thus, it has been found that the terms to be excluded should be defined as stop words. In addition, Uchiyama et al. [8] also performed the process of excluding patent-specific words "**上記,前記,当該,該,毎**" before extracting the terminology.

## 3 DATA AND METHODS

### 3.1 Data

Given that the eighth edition of the IPC has been in use since 2006, we targeted patents filed between January 1, 2006 and December 31, 2019. Here, there were over 4 million patents in all, so it would not be practical to perform an analysis on all of them. Therefore, we narrowed down the FI and ran with it. Because there were still many targets in the section unit (8 kinds), we narrowed the FI down to the class unit. For the class of FI, we referred to information on the revisions of FI published by the Patent Office, and we chose a class that had many revisions (establishment) and not too many overall numbers of revisions (considering the problem of computation time). As a result, out of the top 10 classes that have been newly established since 2006, the lighting class, F21, was chosen as the only one that did not have too many numbers overall. This time, we used only the parts of the summary that we considered to be packed with important information. In total, there were 51,340 patents. We reviewed the yearly progress of the technology.

### 3.2 Flow of This Study

In this study, we performed the analysis with the following stream.

1. Document to Technical Terms (Docs to Technical Terms)
   The pre-processing required to extract Japanese technical terms. On the basis of the results of previous studies, patent-specific words (words used in many patents but not information on technology) are defined as stop words.
2. Dynamic Topic Model
   DTM was used.
3. Evaluation
   For analysis from the new perspective of assisting in the revision of patent classification codes with machine learning, information on the revision of patent classification codes (FI) of the Patent Office was used. Here, we did the same work for the revised information as we did for 1. Then, we extracted the revised technical terms.

---

[‡] Termextract, http://gensen.dl.itc.u-tokyo.ac.jp/termextract.html

[§] "University Research &" http://www.datangraph.com/

Figure 1 shows the whole idea.

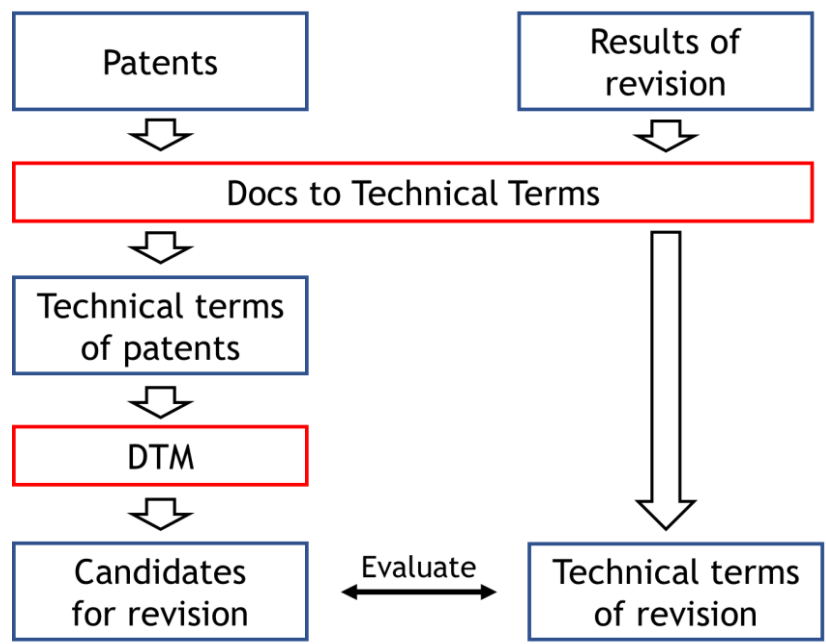

**Figure 1: Flow of This Study**

## 3.3 Details on Methods

### 3.3.1 Docs to Technical Terms

First, we used Python to extract the technical terms. We used MeCab, a library for morphological analysis, and TermExtract, a library for terminology extraction. Next, we defined the stop words in advance. Taking into account the previous studies, the stop words were defined as patent-specific words such as "前記,上記,当該,該当,次,それら,該,毎" and symbols, numbers, and single letters of the English alphabet. In addition, we removed words that appeared frequently in the abstracts (more than 10,000 times for about 50,000 patents) that we considered to be unrelated to the content of the patent classification code. The words deleted here were "summary, task, selected diagram, diagram, solution, plural, invention." Moreover, for the purpose of this study, there was no need to consider less emergent words. Therefore, we focused only on terms that appeared more than 100 times. As a result, 1792 words were extracted as technical terms. When we checked the results, we found that technical terms such as "light-emitting diode" were correctly extracted. (In MeCab, which is often used in Japanese morphological analysis, this would be divided into two words, "light-emitting" and "diode.") In addition, there were 18 patents that did not contain any of the 1792 subject technical terms at all. With the exception of 18 such patents, the total number of patents subject to analysis was 51,322.

### 3.3.2 Dynamic Topic Model

We implemented the dynamic topic model using a Python library, Gensim[**]. We divided the patents into one year each, from 2006 to 2019. Therefore, there were 14 time slices. We set the number of topics to be 20 and the maximum number of iterations until convergence of the Expectation-Maximization algorithm to be 20.

---
[**] https://radimrehurek.com/gensim/index.html

# 4 RESULTS

## 4.1 Results of Experiment

After applying DTM, we could get the technical terms that make up each topic. Table 1 shows the results for topics 1, 2 and 3 with t = 1.

**Table 1: DTM Results for Topics 1, 2, and 3 with t = 1**

| Topic 1 | | Topic 2 | | Topic 3 | |
|---|---|---|---|---|---|
| Technical term | Percentage | Technical term | Percentage | Technical term | Percentage |
| light source | 0.24 | cover | 0.08 | light source device | 0.10 |
| light | 0.22 | case | 0.07 | light-emitting part | 0.08 |
| lighting device | 0.04 | light-emitting body | 0.06 | light-emitting tube | 0.07 |
| incidence surface | 0.03 | opening | 0.05 | reflector | 0.06 |
| projectile surface | 0.03 | LED element | 0.04 | LED chip | 0.04 |

We can see that each topic in DTM was made up of various technical terms. It was found that the topics were divided into those related to light and reflection, those related to fluorescent lamps, and those related to things that emit light, and we could qualitatively confirm that the topics were divided.

Also, in DTM, a patent can be considered to consist of various topics, and soft clustering takes place. The purpose of this study is to assist in the revision of patent classification codes with machine learning. To consider this, we have to take into account the time series of each topic. Therefore, we added up the probabilities of patent affiliation for each time slice to see the transition of each topic. Since the number of patent applications varies from year to year, we normalized sum of probabilities for each year. This value is referred to as the occupation rate.

Figure 2 shows the time series changes for topics 1, 2, and 3.

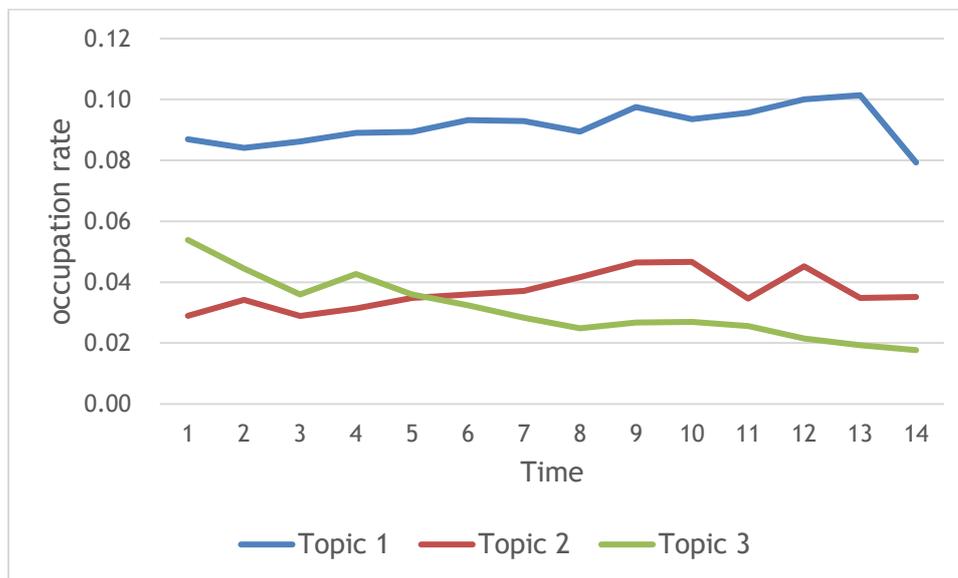

**Figure 2: Time Series Changes for Topics 1, 2, and 3**

Thus, we can see that time series changes occurred for each topic.

In addition, we can see the technical term changes within a topic. Figure 3 shows the temporal evolution of some of the technical terms for Topic 2. We confirmed that the topics themselves have changed over the years.

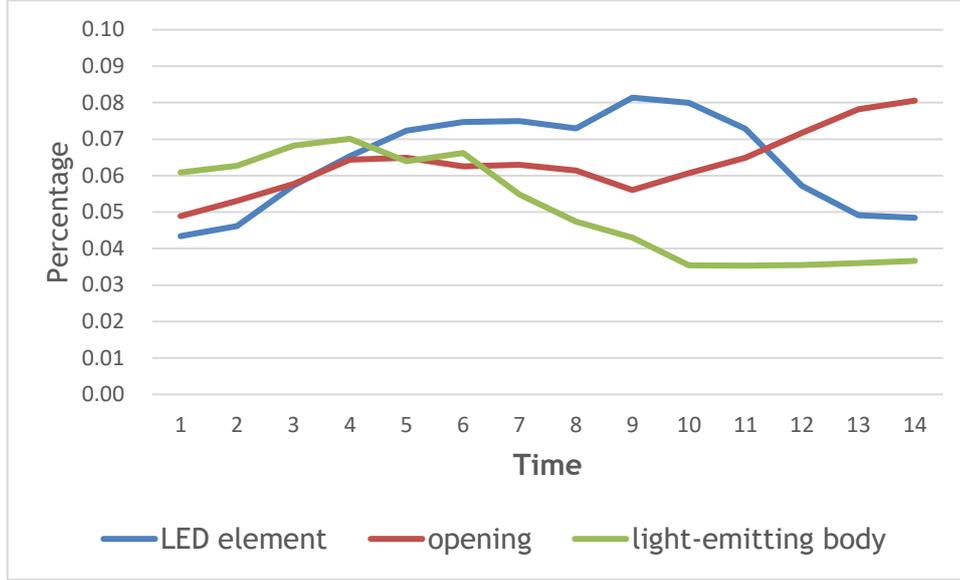

**Figure 3: Temporal Evolution of Some of Technical Terms for Topic 2**

### 4.2 Evaluation

We compared the results with the revision of patent classification codes. For the evaluation, we used information on the revision (establishment) of patent classification codes (FI) from the Patent Office. We performed the same operation as 1 in 3.2 for the established information and extracted the technical terms related to new technology. As a result, a total of 1291 technical terms were extracted. We got the information on the establishment of FI from the Patent Office site[††].

No study has been done to revise the patent classification code, so the metrics were ambiguous. Therefore, we examined indicators. To consider the revision of patent classification codes, how much the technical terms extracted by DTM actually match the terms extracted from the Patent Office data is important. Therefore, we analyzed the experimental results.

We defined the $\text{DTM topic score}(t, j, w)$ as in (1).

$$\text{DTM topic score}(t, j, w) = \text{Percentage of technical term w in topic j at time t}$$

$t \in \{1,2,\ldots,T\}, j \in \{1,2,\ldots,\text{topic\_num}\}, w \in \{\text{the top 100 technical terms in topic j at time t}\}$  *(1)*

In this study, $\text{topic\_num} = 20$ and $T = 14$ because the data are for 14 years. Here, we set N=100. We also calculated a measure of the average occupation rate (Ave topic), as shown in (2), to calculate how much of a topic was dominant. The larger the average share, the more likely it is that the topic is the core of the F21 class. We can get the $\text{DTM topic score}(t, j, w)$ by applying DTM.

$$\text{Ave topic}(j) = \frac{1}{T}\sum_{t=1}^{T} Occupation\ rate(t, j) \quad (2)$$

---

[††] FI revision information: https://www.jpo.go.jp/system/patent/gaiyo/bunrui/fi/f_i_kaisei.html

Then, we introduced the measure of Term score(j, w). This is expressed in equation (3), which shows the percentage of a technical term in the F21 class. An indicator of how much a technical term represents the F21 class can also be produced by the LDA.

$$\text{Term score}(j, w) = \frac{1}{T}\sum_{t=1}^{T} DTM\ topic\ score(t, j, w) \times Ave\ topic(j) \qquad (3)$$

In addition, we checked the time-series change for each technical term of the topic and calculated the amount of increase. For technical terms with a large amount of increase, the probability of revision seems to be high. We normalized the amount of increase. However, even if there is an increase within the topic, if the average share of the topic itself is low, the impact is likely to be small.

Then, we introduced a new measure, which we call the "increase score," which is the normalized amount of increase × the average share of the topic (Ave topic). The larger the measure, the more technical terms there were for the increase as well as the impact on the F21 class. Also, due to the nature of the topic model, a technical term may appear in more than one topic.

In addition, we sorted the technical terms by two scores, the "term score" and "increase score," and we compared them to the actual revision results. Here, we introduce an index used in a recommendation system called AP (Average Precision) that is the average of the fit rate up to and including rank n. AP@n is limited to rank n, and the higher the score, the better the prediction. The maximum value is also 1. Here, the n of @n considered 10, 50, and 100. Table 2 shows the results for all technical patent terms. This suggests that the term score is important.

**Table 2: Results for All Technical Patent Terms**

|  | Sorted by increase score | Sorted by term score |
|---|---|---|
| AP@10 | 0.45 | **1.00** |
| AP@50 | 0.46 | **0.86** |
| AP@100 | 0.45 | **0.80** |

However, terms with a high term score were more likely to be the core technology and known information (e.g., light source, light) of F21. In view of assisting in the revision of patent classification codes with machine learning, it is important to know which words did not fall into the top 100 of Topic j at a certain time but grew rapidly so that they came into the top 100 of Topic j. Therefore, the same analysis was performed on words that were not in the top 100 at a certain time. Table 3 shows the results.

**Table 3: Results for Technical Terms Not in Top 100 at Certain Time**

|  | Sorted by increase score | Sorted by term score |
|---|---|---|
| AP@10 | **0.71** | 0.39 |
| AP@50 | **0.44** | 0.34 |
| AP@100 | **0.35** | 0.21 |

As a result, it is suggested that a measure such as increase score, which can be obtained only from DTM, is more effective than a measure required by LDA for new technologies (e.g., high beam, focusing area) that are easily overlooked by those who revise patent classification codes.

Furthermore, we focused on the technical term "reflector" as an example. Reflector is the technical term for topic 4. Table 4 shows the top results of the term score for Topic 4 and co-occurring revised text.

Table 4: Top Results of Term Score for Topic 4 and Co-occurring Revised Text

| Technical term | Term score | Co-occurring revised text |
|---|---|---|
| one | 0.007 | No match |
| lens | 0.006 | At the focal point, e.g., a refractor, lens, reflector or light source sequence |
| housing | 0.004 | A housing that functions as a reflector |
| two | 0.003 | No match |
| part | 0.002 | Detail of the reflector forming part of the light source |

Thus, we were able to determine that there were technical terms that are actually used simultaneously with reflectors in the revised text. By using the DTM topic model, we could obtain co-occurrence relationships that cannot be understood by simple aggregation.

## 5 DISCUSSION

In the DTM model, the same term appears in different topics in order to allow for flexible expression. Therefore, for some technical terms (light, light source, etc.), there was more than one use for some topics.

F21 is the field of lighting, which is clear as many terms related to light and light sources were used. In the revised data of the patent classification codes of the Patent Office that we used, information such as light and light source also appeared. This is thought to be because they are technical terms related to the core technology of the F21 class, not to new technologies. However, focusing on the limited technical terms, we could identify new technology that were not famous by extracting technical terms that were growing. It helps to revise patent classification codes (FI).

Also, it is possible that there are omissions in the information on the revisions of patent classification codes by the Patent Office, which was used as the correct answer data. Terms may be described in a different way. The difficulty of this study is that it is not known whether the same expression will be used in a revision, which needs to be improved.

## 6 CONCLUSION

In this study, we found topics of patents by DTM after technical term extraction. We then analyzed time-series changes with respect to the topic.

As a result, the technical terms that have an impact on the F21 class were identified as technical terms that are likely to be revised. This is because they are likely to be technical terms that describe the core technology of this class. In addition, we were able to predict new technologies that were not already known by obtaining increasing trends of technical terms from DTM. Furthermore, we found that the results of the topic model revealed the technical words used in combination and were consistent with the actual revision results.

However, since the stop words were set manually, it was found that some technical terms that are considered meaningless were left behind (one, etc.), so it is necessary to consider a mechanism for automatically extracting those that are not technical terms. In addition, by measuring the similarity of terms of revisions of patent classification codes using Word2Vec, etc., it is necessary to solve the problem that the expressions used in patents are slightly different from the technical terms of revisions of patent classification codes.